\def\year{2021}\relax
\documentclass[letterpaper]{article} 
\usepackage{aaai21}  
\usepackage{times}  
\usepackage{helvet} 
\usepackage{courier}  
\usepackage[hyphens]{url}  
\usepackage{graphicx} 
\usepackage{natbib}  
\usepackage{caption} 
\usepackage{subfigure}
\usepackage{multirow}
\usepackage{threeparttable}
\usepackage{booktabs}

\usepackage{color}

\title{Team Resilience under Shock: An Empirical Analysis of GitHub Repositories during Early COVID-19 Pandemic }
\author{
    Xuan Lu,\textsuperscript{\rm 1}
    Wei Ai,\textsuperscript{\rm 2}
    Yixin Wang,\textsuperscript{\rm 3}
    Qiaozhu Mei\textsuperscript{\rm 1}
    \\
}
\affiliations{
    \textsuperscript{\rm 1}School of Information, University of Michigan, Ann Arbor, Michigan, USA\\
    \textsuperscript{\rm 2}College of Information Studies, University of Maryland, College Park, Maryland, USA\\
    \textsuperscript{\rm 3}Department of Statistics, University of Michigan, Ann Arbor, Michigan, USA

    luxuan@umich.edu, aiwei@umd.edu, yixinw@umich.edu, qmei@umich.edu

}

\begin{document}

\maketitle

\begin{abstract}
While many organizations have shifted to working remotely during the COVID-19 pandemic, how the remote workforce and the remote teams are influenced by and would respond to this and future shocks remain largely unknown. Software developers have relied on remote collaborations long before the pandemic, working in virtual teams (GitHub repositories). The dynamics of these repositories through the pandemic provide a unique opportunity to understand how remote teams react under shock. This work presents a systematic analysis. 

We measure the overall effect of the early pandemic on public GitHub repositories by comparing their sizes and productivity with the counterfactual outcomes forecasted as if there were no pandemic. We find that the productivity level and the number of active members of these teams vary significantly during different periods of the pandemic. We then conduct a finer-grained investigation and study the heterogeneous effects of the shock on individual teams. We find that the resilience of a team is highly correlated to certain properties of the team before the pandemic. Through a bootstrapped regression analysis, we reveal which types of teams are robust or fragile to the shock. 

\end{abstract}
\section{Introduction}

The COVID-19 pandemic has been a shock to almost every aspect of society, not sparing the productivity and sustainability of organizations. As a response to the pandemic, many organizations have temporarily (some even permanently) shifted to working remotely, a paradigm of work that depends much less on centralized workplaces or in-person collaborations, and a paradigm that has long been considered as a potential trend of the future of work. The pandemic shock provides a surprise ``test drive'' and a unique opportunity to collect many first-hand observations about the behavior of remote workers, especially in a \textit{team} setting that normally requires lots of in-person interactions. Indeed, the impacts of the shift  to remote work have been noted on teamwork and collaborations~\cite{yang2022effects}. However, as most teams are recently ``switched'' to remote work (due to the pandemic), there is a strong confound that hinders our understanding of how these teams would have behaved if they had been used to working remotely. 

These are critical questions for organizations and individuals when they make decisions between continuing remote teamwork or switching back to in-person teamwork as the pandemic dissolves (hopefully!) and in the further future. For example, how would remote teams be influenced by similar external shocks in the future, if they had truly become the norm? How quickly can they bounce back?  What kind of remote teams are more resilient under shock? Answers to these questions remain largely unknown.

Answering such questions requires estimating the causal effects of the pandemic on remote teams, where the pandemic shock can be viewed as the ``treatment'' in a quasi-experiment setting. To conduct such an analysis, one needs to observe the behavior of remote teams before and after the treatment, which unfortunately rules out the teams that were not remote before the pandemic. Fortunately, there are teams that have relied on online collaborations long before the pandemic, such as developers who work in the same repositories on GitHub, the leading online collaborative platform for software development. The GitHub repositories are often created by companies (especially in the IT industry), voluntary open source software (OSS) contributors, non-profit organizations, and researchers. These teams have been collaborating remotely way before the pandemic, and they provide natural observations both before and after the shock. Therefore, the GitHub teams provide a unique testbed to answer these research questions, even though the answers may be interpreted in the context of open-source software (OSS) communities and related organizational scenarios. 

However, as all the teams are exposed to the pandemic shock (the treatment), it is challenging to estimate the counterfactual outcomes (what if they were not treated) due to the lack of a control group. To address this challenge, we propose an approach close to the interrupted time series analysis~\cite{mcdowall2019interrupted} to make predictions about the counterfactual outcomes. Specifically, we train machine learning models to predict the outcome of a remote team based on its properties and historical behavior, as if the shock had never happened, and use the predictions as the estimates of the counterfactual outcome. The individual treatment effect (ITE) of the shock on a team is then estimated as the difference between the observed outcome after the shock and the estimated counterfactual outcome. To reveal what kind of teams are more resilient to the shock, we analyze the relations between team properties and the ITE through bootstrapped regressions that take into account the distribution of errors of the counterfactual prediction, and we derive findings that are robust to the potential errors. 

We conduct our analysis on the public GitHub repositories (as surrogates of virtual, remote teams) before and during the early months of the COVID-19 pandemic, based on a longitudinal dataset of event logs collected by GHArchive\footnote{http://www.gharchive.org/, retrieved on March 18, 2021.} and the profile information of users in these repositories\footnote{https://ghtorrent.org/, retrieved on May 8, 2022.} collected by GHTorrent~\cite{Gousi13}. To identify the overall shock effect on GitHub teams, we compare the observed activity levels of these repositories and the expected activity levels obtained through time series forecasting. Then, we examine the shock effects on individual teams more rigorously. We design a comprehensive set of features to characterize the properties of a team. Using advanced machine learning models trained with these features, we are able to predict the expected status and activity levels of these teams after the shock, which helps us estimate its effects on individual teams and analyze the heterogeneity in the effects.  

Our main contributions can be summarized as follows:

\begin{itemize}
    \item We study the general effect of the pandemic shock on remote teams (using GitHub repositories as examples) through time series forecasting and demonstrate the effect with visual examination.
    \item We find that team productivity was immediately reduced when the pandemic hits, while team growth (measured by active members) responds to the shock in a longer term.
    \item We estimate per-team treatment effects of the pandemic through machine-learning based counterfactual predictions, and we identify team properties that are significantly associated with team resilience under the shock through bootstrapped regressions that are robust to the potential errors of the counterfactual predictions.
\end{itemize}
\section{Related Work}
\subsection{GitHub Repositories}

Using GitHub repositories as surrogates for teams is common practice in literature~\cite{maldeniya2020herding, ortu2017diverse, Vasilescu2015-ql, Jarczyk2014
}. For example, \cite{maldeniya2020herding} studies how repositories as virtual teams respond to increased attention. \cite{Vasilescu2015-ql} investigates how tenure and gender diversity of developers relate to team productivity and turnover. Technique details in team selection can be different due to the complexity of data~\cite{Kalliamvakou2014-rr} and distinct research objectives. We limit the number of active members and contribution histories of repositories to filter teams where members work remotely and consistently in Section~\ref{subsec:selection}.

The performance of GitHub repositories can be measured in different aspects: productivity~\cite{Vasilescu2015-ql, meyer2014software, vasilescu2015quality, mockus2002two}, team size~\cite{mcdonald2013performance}, code quality~\cite{mockus2002two}, etc.. The productivity of a repository is also a combination of different activities. It could be measured by the quantity of the activities such as \textit{commits}~\cite{Vasilescu2015-ql} or \textit{pull requests}~\cite{github2020spotlighht}, time spent on code~\cite{meyer2014software}, accepted changes \cite{vasilescu2015quality} (i.e., the number of merged \textit{pull requests}), file changes~\cite{casalnuovo2015developer}, or the response time to problem reports submitted by users~\cite{mockus2002two}. We select the number of \textit{pushes} as a metric of productivity, following the official report of GitHub~\cite{github2020spotlighht}, and the number of active members as an additional metric to characterize the growth of a repository during the pandemic.

A cluster of work has been done to explore the properties related to the performance of virtual teams (and GitHub repositories in particular), including the demographic diversity in gender~\cite{Vasilescu2015-ql}, country~\cite{ortu2017diverse}, languages spoken~\cite{daniel2013effects}, experiences and expertise~\cite{Vasilescu2015-ql}, social connections~\cite{casalnuovo2015developer, blincoe2016understanding}, the ability of multitasking~\cite{vasilescu2016sky}, and the emotional status~\cite{graziotin2014happy} of developers. We propose a comprehensive set of features that characterizes many of these properties in Section~\ref{subsec:property} to understand team resilience under the shock.

\subsection{Shock Effect Estimation}

Since the pandemic can be seen as an external shock out of the developers' control, we adopt a causal inference framework to estimate the effect on each team. There are several causal inference frameworks developed to evaluate the impact of a shock, they don't fit this problem very well. Difference-in-Differences models \cite{angrist2008mostly,ye2020predicting} are commonly used to analyze ``natural experiments,'' yet the pandemic influences the entire GitHub community, leaving no control group to construct the counterfactuals. Regression Discontinuity in Time \cite{hausman2018regression} has been proposed to evaluate the impact of a policy that universally affect all subjects at a cutoff time. However, as with all Regression Discontinuity methods, the estimated treatment effect is usually focused on the narrow window close to the cutoff. The shock from the pandemic is much more complex, because both the infection rate and the workspace policies are rapidly changing over time, and we need to estimate the treatment effect over a longer time span. To this end, our approach is mostly close to the interrupted time series analysis \cite{mcdowall2019interrupted}. Instead of fitting linear models for forecasting, we adopt advanced machine learning algorithms to predict the counterfactuals in Section~\ref{sec:effect}.  
\section{GitHub Repositories under the Pandemic}\label{sec:time_series}
As a leading online collaborative platform for open-source software development, GitHub has been used by millions of developers for years. Developers who actively contribute to the same repositories should be already familiar with remote collaborations. Are these remote teams influenced by the shock of a worldwide pandemic? 

\subsection{GitHub Repositories}
We base our analysis on the event log data archived by GHArchive\footnote{http://www.gharchive.org/, retrieved on March 18, 2021.} from January 2015 to December 2020. The dataset includes more than 20 types of events provided by GitHub, attributed to individual developers and \textit{public} repositories.\footnote{https://developer.github.com/webhooks/event-payloads/, retrieved on October 9, 2021.} 
Although the events correspond to the activities of individuals (code submission, issue tracking, coordination, etc.), a few event types should not be counted towards their contributions to a team, such as a \textit{watch} event. A developer is not an active contributor to a repository if they simply ``watch'' it without performing other activities. Similarly, when a developer \textit{forks} a repository, they are not contributing to the original repository unless they merge the code back in. Therefore, we exclude \textit{watch} and \textit{fork} and use all other event types to measure the activeness of repositories.\footnote{\textit{Star} events also belong to the excluded category but they are not included in our dataset.} In a given period of time, a repository is regarded as active if there is at least one logged event. A developer is regarded as an active member of a repository if they have at least one logged event explicitly attributed to this repository. In this way, we obtain a dataset containing 27.6 million users that have been active contributors to at least one repository and 160.5 million repositories that have been contributed by at least one member during the six years.

We first count the number of active repositories and the number of active developers per repository on a monthly basis. The latter provides a measurement of the effective size of the team in that month.  Furthermore, we measure the productivity of a repository by counting two events directly related to code contribution~\cite{github2020spotlighht}. Specifically, \textit{pushes} are used to upload all local branch commits to the corresponding remote branch; \textit{pull requests} are used to tell others about changes pushed to a branch with the aim of getting the changes merged to the \textit{base} branch. In our dataset, there are 115 million opened \textit{pull requests} and 1.5 billion \textit{pushes} in total. We aggregate these statistics over active repositories every month to measure the monthly productivity of the GitHub platform. 

\subsection{The Pandemic Shock}

Intuitively, the productivity of both individual teams and the platform as a whole may have been impacted when the pandemic hit. Since it is a worldwide pandemic, it is hard to define the exact date of the shock or to pin down the dates to individual developers or repositories even if one could link the identities of developers. Indeed, the behavior of a team may be already influenced even before the wave hits their own locations. Considering that COVID-19 first outbroke in late January of 2020, we regard the whole year of 2020 as possibly affected by the pandemic when estimating the shock effects. In particular, the first three months of 2020 overlap with the first outbreaks of COVID-19 and the declaration of the pandemic by the WHO (on March 11, 2020),\footnote{\url{https://www.who.int/director-general/speeches/detail/who-director-general-s-opening-remarks-at-the-media-briefing-on-covid-19---11-march-2020}, retrieved on January 15, 2021.} which reflect the immediate shock effect of the pandemic. The months shortly after (i.e., April to June) are more reflective of the teams' responses to remote work, as when the pandemic unfolds, working from home has been adopted as a common response in most countries. The months thereafter are more reflective of their longer-term responses to the pandemic, and by then remote work should have been widely adopted and COVID vaccines had not been deployed. 

\subsection{Overall Effects of the Shock}

We anticipate that the influence of the pandemic on GitHub teams may vary depending on the stages. For each of the aggregated productivity metrics, we construct a time series with monthly measurements since January 2015 and use the observations before 2020 (60 months in total) to forecast the values in the next 12 months (i.e., January-December 2020) with an exponential smoothing and STL decomposition method~\cite{robert1990stl}. 

\begin{figure*}[htb]
    \centering
    \begin{center}
        \subfigure[Active repositories \label{fig:sub_repo_12}]
        {\includegraphics[width=0.245\linewidth]{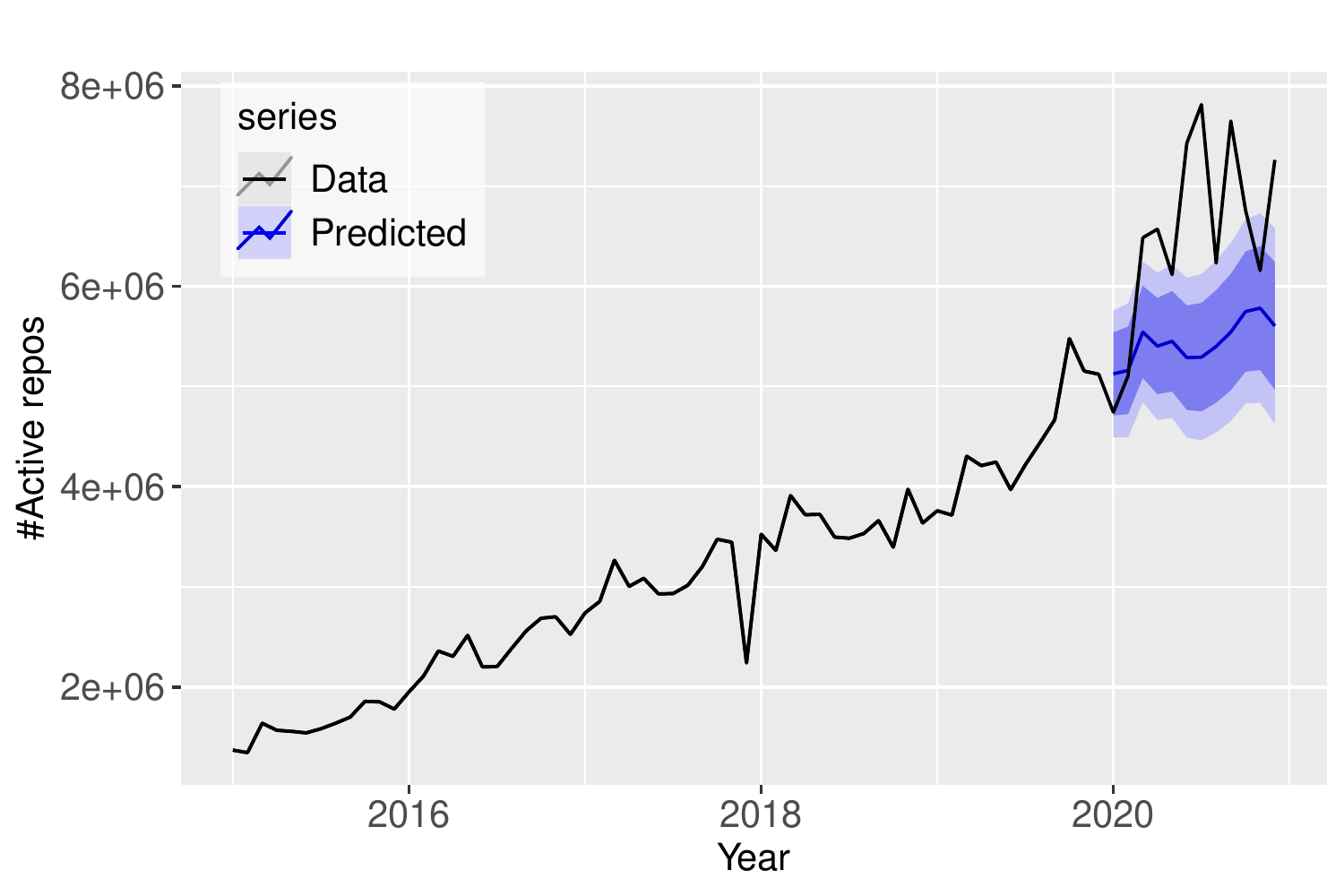}}
        \subfigure[Opened pull requests\label{fig:sub_pr_12}]
        {\includegraphics[width=0.245\linewidth]{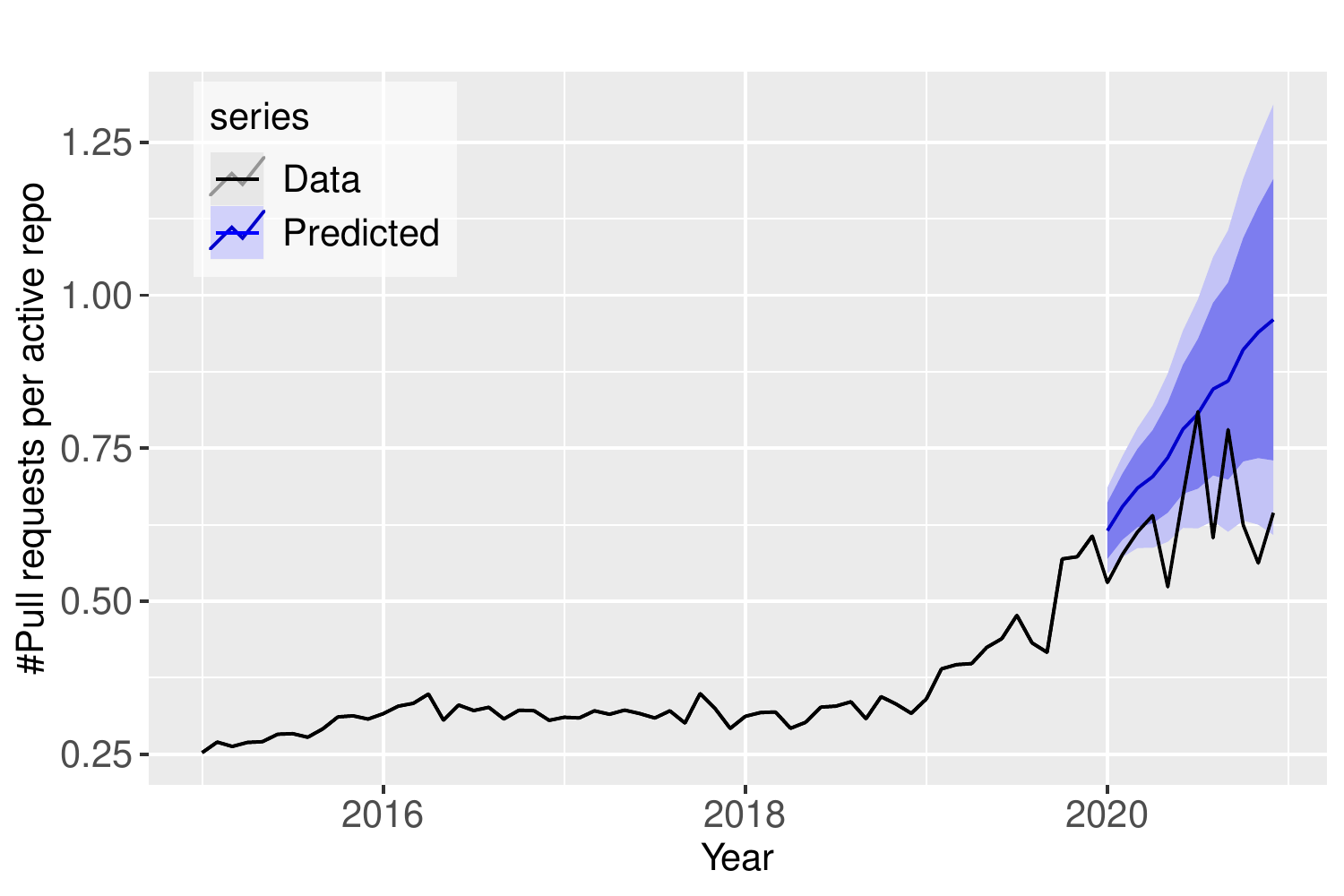}}
        \subfigure[Active team size \label{fig:sub_actor_12}]
        {\includegraphics[width=0.245\linewidth]{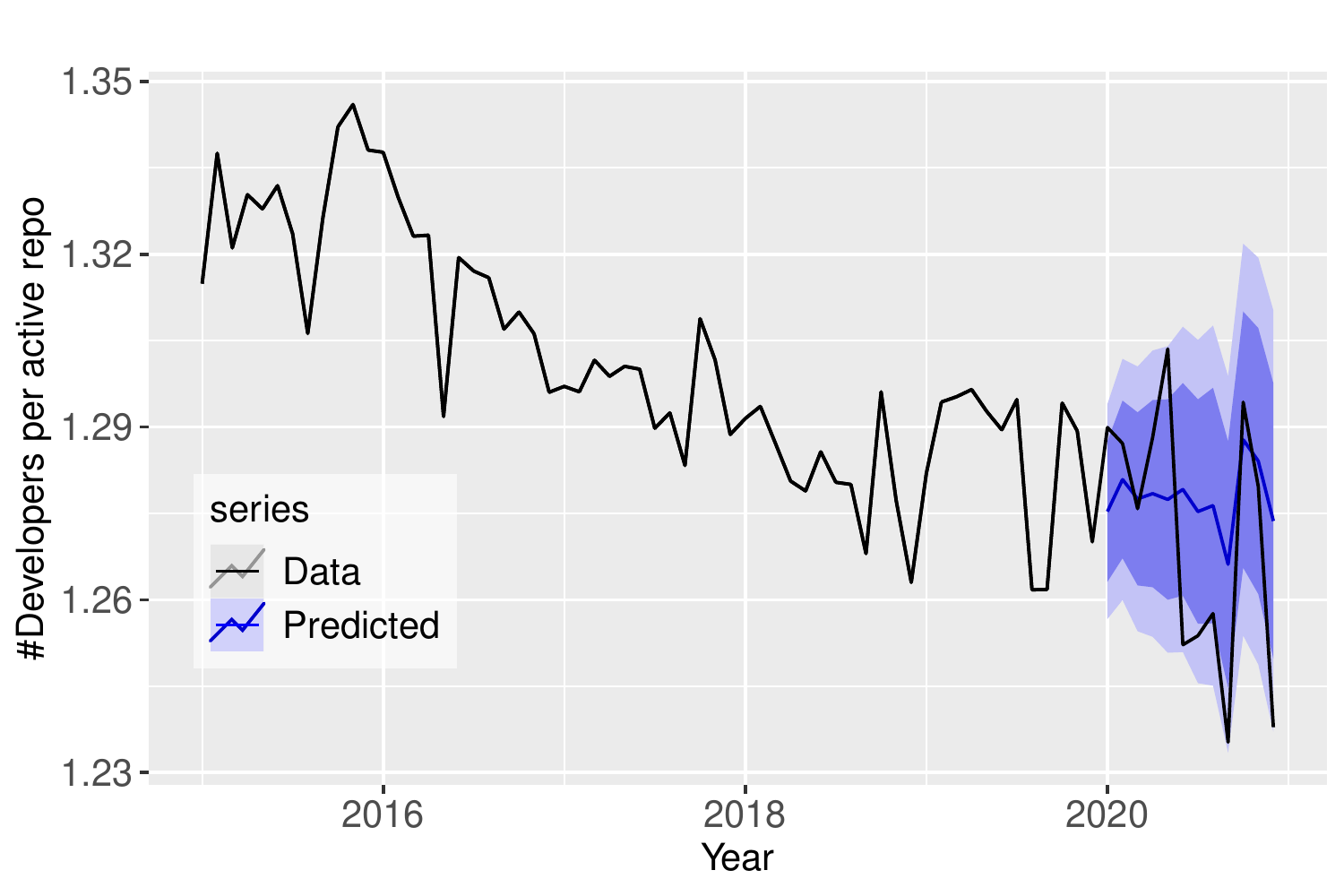}}
        \subfigure[Pushes \label{fig:sub_push_12}]
        {\includegraphics[width=0.245\linewidth]{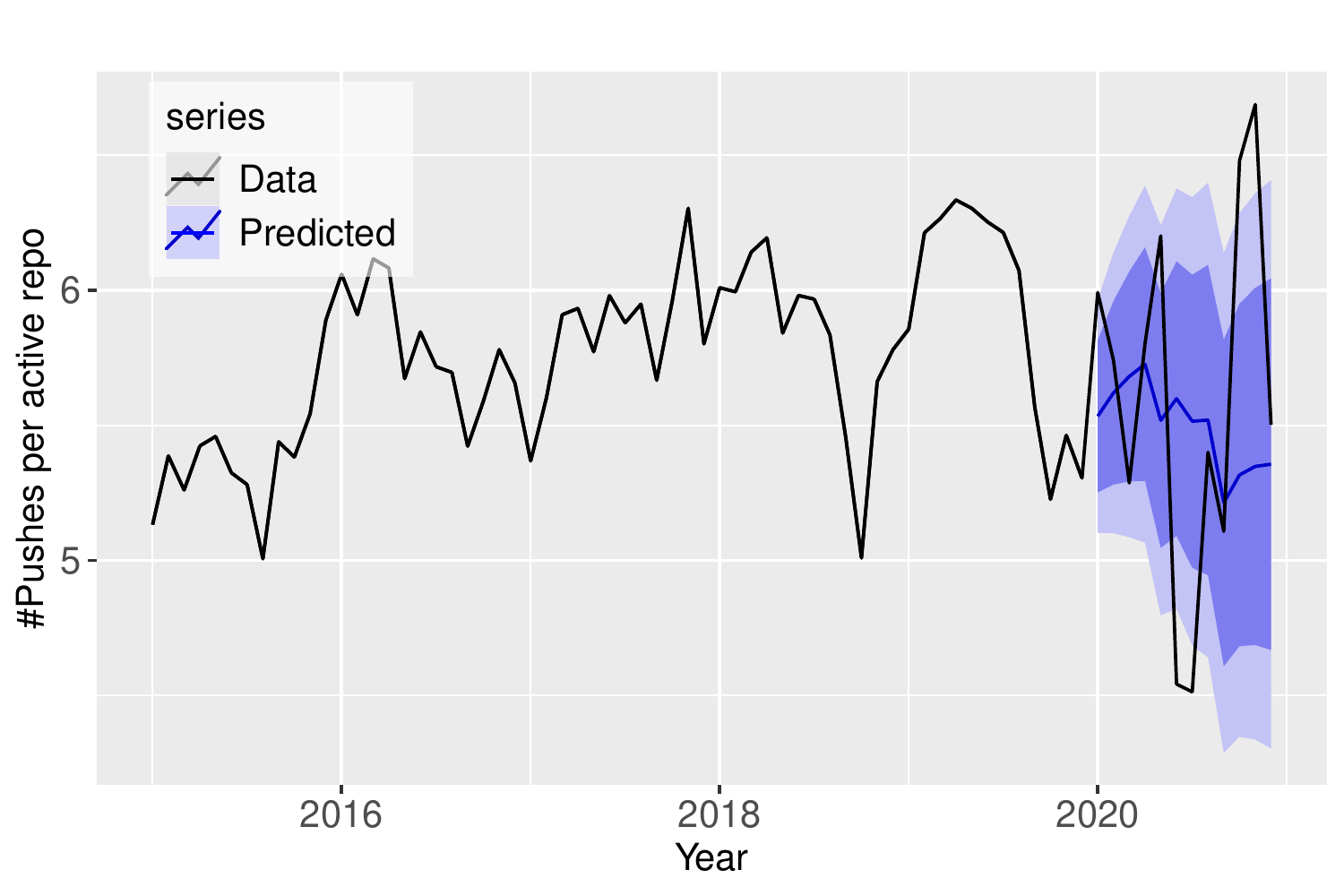}}
        \caption{Monthly status of repositories from 01/2015 to 12/2020 compared with forecasted status from 01/2020 to 12/2020. Deep blue: 80\% confidence interval; light blue: 95\% confidence interval. The pandemic shock has a \textbf{positive} effect on the number of active remote teams, a \textbf{negative} effect on pull request activities, and a \textbf{mixed} effect on team size and push activities. }\label{fig:time_series_12}
    \end{center}
\end{figure*}

Figure~\ref{fig:time_series_12} visualizes the observed values of four metrics compared with their forecasted values (what if there were no pandemic). The differences can be interpreted as the overall effects of the pandemic on the GitHub repositories. As shown in Figure~\ref{fig:sub_repo_12}, the number of \textit{active repositories} shows an increasing trend over years, and the predicted number of active repositories during 2020 follows the same trend. Interestingly, the observed numbers of active repositories after the pandemic started are significantly larger than expected since March 2020 when COVID-19 becomes a worldwide pandemic, which shows that the shock does have an effect on the activeness of the repositories. It is likely that many new teams start to use GitHub, or existing inactive teams have reactivated their GitHub repositories for remote collaborations. This finding is aligned with the trend that the pandemic has made teams shift to remote collaborations, and GitHub provides an efficient platform for this shift. 

Although the number of active teams has increased because of the pandemic, the other measures show different patterns. For example, Figure~\ref{fig:sub_pr_12} shows that although there is an increasing trend in the number of \textit{opened pull requests} before the pandemic and it is projected to further increase in 2020, the actual observations are mostly below the expectation. This indicates that while more teams have joined (or rejoined) GitHub, the productivity of existing teams may have actually dropped. An alternative possibility is that the newly activated teams might be just smaller (and thus less productive). However, Figure~\ref{fig:sub_actor_12} does not suggest a clear trend of the effect of the pandemic on the size of active teams, other than it becomes much more volatile during the pandemic. The number of active members of a repository actually increased during the first five months (although not significantly) and then dropped below the forecasts.   
The pandemic effects on the number of \textit{pushes} per repository present similar patterns as team size, although in certain months the difference between observed and forecasted are significant.

The different patterns on \textit{pushes} and \textit{pull requests} are rather interesting.  While both measure productivity, the former is more of individual activity, thus more correlated to the number of members in a team; the latter represents a collaborative activity (someone else in the team is expected to merge the updates into the base branch). The visual examination suggests heterogeneity in the shock effects over time, on different outcomes, and potentially on different types of teams. This motivates us to conduct a more rigorous analysis to examine the shock effect on individual repositories.

\section{Shock Effects on Individual Repositories}\label{sec:effect}

The preceding analysis has shown the overall effects of the pandemic on all GitHub repositories. If we consider the pandemic shock as an intervention, these effects are similar to the notion of ``average treatment effects (ATE)'' in causal inference literature.  The difference is that all the teams are exposed to the pandemic (they are all ``treated''), therefore there isn't a control group of teams that are not ``treated'' by the pandemic so we can use it to estimate the counterfactual outcome. Alternatively, the counterfactual outcome is estimated through time series forecasting.  The average effects of the shock have a high variance among individual repositories. We seek to zoom in on the analysis and understand how the pandemic is affecting individual repositories/teams.

Formally, the ``individual treatment effect'' (ITE) of the shock on a team is the difference between the \emph{potential outcome} of the team if we made the team exposed to the pandemic shock and the \emph{potential outcome} if we made the team not exposed to the shock. Under the ignorability assumption---namely the treatment (i.e. exposure to the pandemic shock) is ignorable given all the team information before the shock---the ITE estimated by the difference between the \textit{observed} outcome of the team shortly after the shock and the \textit{expected} outcome if the shock had never happened~\citep{xu2017generalized}. That is, the ITE of the repository $j$ at the time $t$ can be estimated as

\begin{equation}
\label{eq:ite}
\widehat{ITE}_{j,t} = Y_{j,t} - \hat{Y}_{j,t},
\end{equation}
where $Y_{j, t}$ is the observed outcome (e.g., productivity or size) of team $j$ at time $t$ (after the pandemic started) and $\hat{Y}_{j,t}$ is the predicted outcome based on team information before the pandemic shock. 

The outcome of a team ($Y$) can be measured in different ways. We are particularly interested in the productivity and growth of the team.  We select the team size (the number of active members) as a metric of growth and the number of \textit{pushes} as a metric of productivity.  The selection of $t$ can be either short-term or longer-term. We do not include a longer horizon over 6 months with the concern that it may introduce many more confounds (evidenced by the high volatility among the later months in Figure~\ref{fig:sub_actor_12} and ~\ref{fig:sub_push_12}). 

An important aspect of Equation~\ref{eq:ite} is the estimation of counterfactual $\hat{Y}_{j,t}$. In conventional causal inference literature, this is usually done by matching treated subjects with untreated subjects that have similar properties. In our context, as every team is ``treated'' by the pandemic, we take a practical approach by making a forecast of $\hat{Y}_{j,t}$ based on the status of the team before the pandemic. This is similar to what we did for the aggregated time series in Section~\ref{sec:time_series}, but this time we could consider many more properties of the individual teams to forecast their counterfactual outcome, through the help of a machine learning predictor.  
In particular, we train a machine learning model that uses the properties of each repository in the 4th quarter (Q4) of 2018 to predict their outcomes ($Y_{j, t}$) of a specific month in 2019. We then use the trained model to forecast the outcome of repositories of the same month in 2020 ($\hat{Y}_{j, t}$) based on their features in Q4 of 2019. Note that these counterfactual models also provide an understanding of which teams are more productive and grow faster when they are \textit{not} under shock. 

Note that in theory, we could re-estimate the average treatment effects of the pandemic shock in Section \ref{sec:time_series} by computing the mean of the individual treatment effects over all treated teams (i.e., $\frac{1}{N}\sum_{j = 1..N}\widehat{ITE}_{j,t}$), in practice this can be problematic as the predicted outcome $\hat{Y}_{j,t}$ may be inaccurate for a large number of teams that have limited information before the pandemic. This is in line with our concerns that if most teams are recently switched to remote work, it is hard to find out how they would have behaved if they had always been remote. To remove this confound, we analyze the individual treatment effect on a carefully selected set of teams that are already active before the pandemic and we can trust their counterfactual predictions. 

\subsection{Team Selection}\label{subsec:selection}
We define \textbf{team members} as those who make any type of contribution to the repository, following the survey findings in~\cite{Vasilescu2015-ql}. As stated before, developers who are just \textit{watching} or \textit{forking} a repository are excluded.

We select two sets of repositories in this analysis. The \textbf{\textit{reference set}} contains 71,928 repositories that were active in \textit{every} quarter of 2018 with at least \textit{three} active members. This excludes teams that were built temporarily for short-term projects (such as course teams). These repositories must have logged \textit{push} events by 2018. Similarly, there are 87,914 repositories in the \textbf{\textit{target set}} had at least three active members in \textit{every} quarter of 2019 and push records by 2019. The criteria are set to identify consistently maintained repositories that are already familiar with remote work.

\subsection{Team Properties}\label{subsec:property}
For repositories in the reference set and target set, we characterize their team properties in the 4th quarter of their selected year (i.e., 2018/2019 for the reference/target set) in the following aspects, respectively.\footnote {We use the GHTorrent Mysql dump dated 2021-03-06 retrieved from \url{http://ghtorrent-downloads.ewi.tudelft.nl/mysql/} to extract the metadata of users (account creation time and country) and their followers. Other information is from the GHArchive data.}

\noindent $\bullet$ \textbf{Team size}. This feature directly measures the number of active members in a team, which could be associated with productivity levels and future team sizes. 

\noindent $\bullet$ \textbf{Multitasking}. Developers could be a member of multiple repositories, while frequently switching between different contexts can lead to distraction or affect productivity~\cite{vasilescu2016sky}. For each team, we count the members that work for only this team and the average number of repositories contributed by all members.

\noindent $\bullet$ \textbf{Tenure}. Tenure refers to the experience of a member, which could be measured from different aspects. Following~\cite{Vasilescu2015-ql}, we measure three types of tenure for each developer. The \textit{platform tenure} captures one's presence on GitHub since they created the account. The \textit{coding tenure} measures one's coding experience by locating their first \textit{push} on GitHub. The \textit{team tenure} is the days since one made their first contribution to the team. To characterize the tenure distribution of team members, we adopt the maximum, median, and diversity statistics of each tenure measure. The diversity is measured by both the standard deviation and the coefficient of variance (CV)~\cite{Vasilescu2015-ql}, which is defined as the ratio of the adjusted standard deviation to the mean of a list of numbers. 

\noindent $\bullet$ \textbf{Prestige}. We measure the prestige of team members with their numbers of followers, which could also be an indicator of social influence~\cite{ortu2017diverse}. Similarly, to characterize the distribution of member prestige, we use the maximum, median, and diversity of the values of a team. 

\noindent $\bullet$ \textbf{Demographics}. We use the countries from their self-reported profiles as an example of the demographics of team members. The number of countries and the entropy~\cite{shannon2001mathematical} of the country distribution are used to characterize the country diversity in a team. Note that the country information could be absent, we compute the features with team members whose country information is disclosed.

\noindent $\bullet$ \textbf{Programming languages}. Programming language is regarded as a salient and noticeable identity of a developer on GitHub. For each team, we measure the number of programming languages that GitHub developers mostly use and the entropy of the programming language distribution.

\begin{figure}
\begin{center}
    \includegraphics[width=0.85\linewidth]{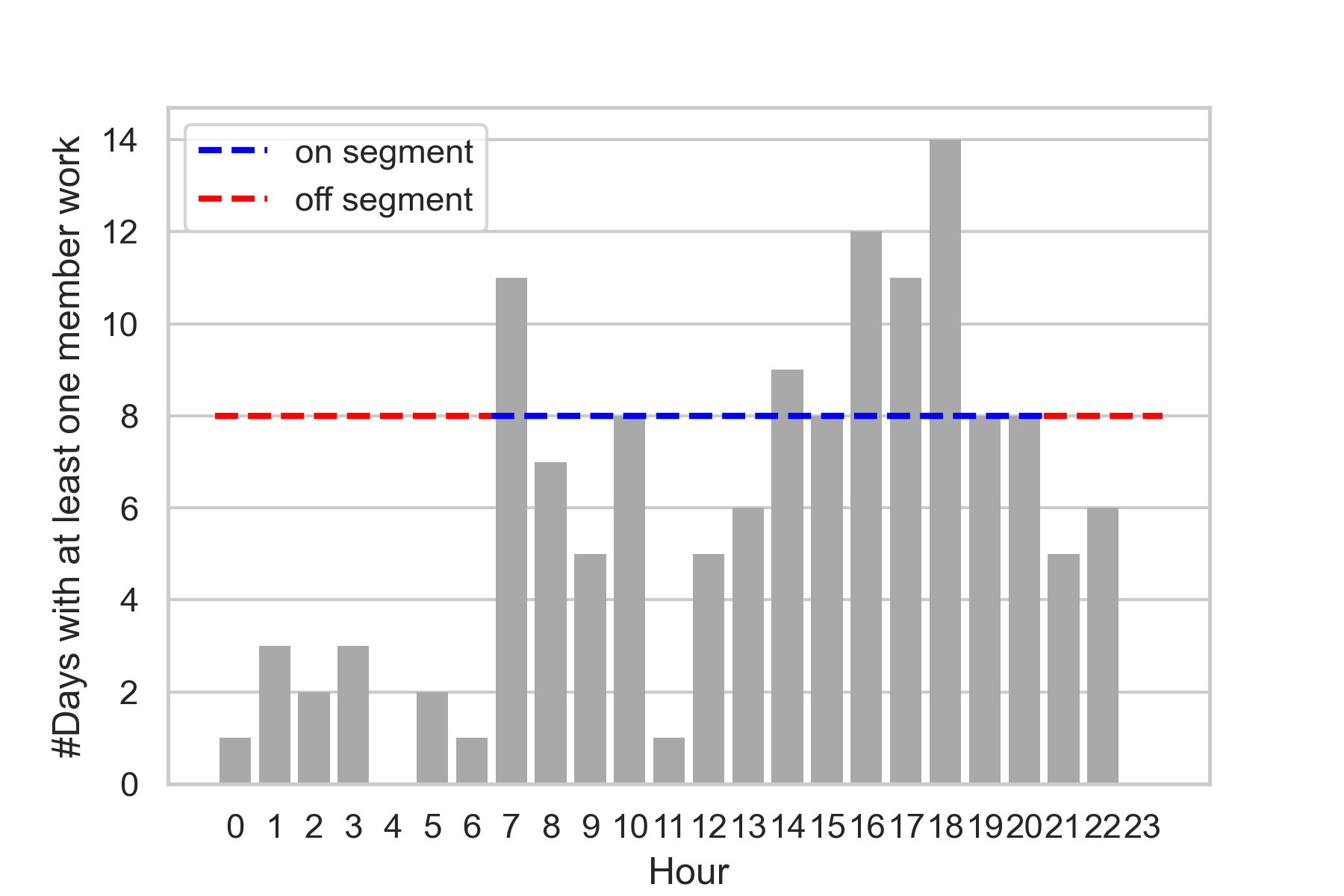}
    \caption{An example of the off segment of a repository. The bar plot shows the number of days (in a quarter) that team members worked in a given hour of a day. The off segment characterizes the longest consecutive hours that people worked for less than a certain number of days (e.g., 8). The off segment of the shown repository starts at 21:00 and ends before 7:00, which indicates a 10-hour-long off segment.} 
    \label{fig:off_segment_example}
    \end{center}
\end{figure}

\noindent $\bullet$ \textbf{Working schedule}. Because the dataset does not provide the time zones of users, we measure the work schedule of members in a team with the entropy of their working hours over 24 hours of a day. Specifically, we construct a 24-dimensional vector to describe the number of days in a quarter with member(s) working in each hour. The entropy of hours describes the distribution of this vector. A higher entropy indicates more spread-out working hours, while a lower entropy could indicate clear boundaries between work and life. Furthermore, we determine the \textit{off segment} of a team from this vector to describe the routineness of the team's working schedule. Figure~\ref{fig:off_segment_example} illustrates an example.

Meanwhile, considering that the activity and outcome of developers show a weekly pattern~\cite{github2020spotlighht}, we also construct a 7-dimensional vector to describe the distribution of working days in the 7 days of a week. Entropy is used to characterize this distribution.

\noindent $\bullet$ \textbf{Communication}. We use the number of comments to \textit{issues}, \textit{pull requests}, or \textit{commits} to reflect the communications of a team~\cite{Vasilescu2015-ql}. As the emotion status of developers is correlated with productivity~\cite{graziotin2014happy}, we adopt emojis as a sensor of emotion expression. The proportion of emojis in the communications posts is used to characterize this property. 

\noindent $\bullet$ \textbf{Repository age}. The age of a team could be related to its outcome. We compute the age for a repository with the days between its creation and the last day of the selected year.

Additionally, we include the following variables to control for the confounds of the team outcome during the shock. We measure the number of \textit{pushes}, \textit{pull requests}, and \textit{issues} in the 4th quarter of and the monthly \textit{pushes} by the selected year. Literature has shown that external attention could influence team behaviors~\cite{maldeniya2020herding}, to control for this, we count the overall number of \textit{watches} and \textit{forks} by the selected year and the recent attentions in its 4th quarter.

\subsection{Prediction Models}
With the properties of the reference repositories in the 4th quarter of 2018, we then train machine learning models to predict their outcomes for a given month in 2019. Specifically, we train one model for each of the six selected months (i.e., January-June, denoted as month $1$-$6$) to predict each of the two outcomes (i.e., productivity and growth).

We adopt a Gradient Boosting Decision Tree model (GBDT) and a Random Forest model (RF) for this prediction, implemented with scikit-learn~\cite{scikit-learn}. After dropping repositories with missing features, we have 42,018 repositories in the reference set. We use an 80-20 train-test split for this set, and tune hyper-parameters on training with 5-fold cross-validation. To compare, we use a seasonal na\"ive model as the baseline, which predicts the outcome of a repository in month $i$ of 2019 with the outcome of the same repository in month $i$ of 2018. 

We train a model for each of the month $i$ ($i\in\{1, 2, .., 6\}$) and present the performance of the models in Table~\ref{tab_prediction_2}. Results show that both the GBDT and RF models obtain a much higher $R^{2}$ and a much lower $MSE$ than the baseline seasonal na\"ive model across all six months. Similar observations could be made from the prediction models for team size. As GBDT performs the best in most settings, we adopt it in the rest of the paper. The $R^{2}$ scores are mostly $>$0.6 and even higher for a shorter time horizon, which is reasonable given the difficulty of the prediction task.

Note that a high $R^{2}$ (or a low $MSE$) does not ensure the robustness of downstream analyses, which may still be affected by the potential errors in the counterfactual predictions. These errors are not observable and thus cannot be directly measured. To address this issue, we generate a reasonable estimation of the unobserved errors of the counterfactual prediction in Section~\ref{sub:conformal}, which can be used to analyze the potential effect of these errors on downstream analyses.

\begin{table*}[]
\small
\caption{Performance of counterfactual predictions (on first six months $i$ of 2019). GBDT performs the best in most settings. }
\begin{tabular}{ccccccccccccccc}
\toprule
 & \multicolumn{1}{l}{} & \multicolumn{6}{c}{$R^2$} & \multicolumn{1}{l}{} & \multicolumn{6}{c}{MSE} \\\cmidrule{3-8} \cmidrule{10-15} 
 & & $i=1$ & $i=2$ & $i=3$ & $i=4$ & $i=5$ & $i=6$ & & $i=1$ & $i=2$ & $i=3$ & $i=4$ & $i=5$ & $i=6$ \\\cmidrule{1-8} \cmidrule{10-15} 
\multirow{3}{*}{Productivity} & GBDT & \textbf{0.652} & \textbf{0.614} & \textbf{0.595} & \textbf{0.575} & \textbf{0.549} & \textbf{0.519} & & \textbf{1.727} & \textbf{1.994} & \textbf{2.171} & \textbf{2.247} & \textbf{2.428} & \textbf{2.485} \\
\multicolumn{1}{l}{} & RF & 0.647 & 0.607 & 0.588 & 0.569 & 0.541 & 0.511 & & 1.752 & 2.032 & 2.209 & 2.276 & 2.468 & 2.525 \\
 & Baseline & 0.094 & 0.155 & 0.182 & 0.213 & 0.210 & 0.190 & & 4.493 & 4.365 & 4.391 & 4.155 & 4.251 & 4.180 \\\midrule
\multirow{3}{*}{Active Team Size} & GBDT & \textbf{0.711} & \textbf{0.678} & \textbf{0.671} & \textbf{0.636} & \textbf{0.638} & 0.604 & & \textbf{0.592} & \textbf{0.686} & \textbf{0.729} & \textbf{0.600} & \textbf{0.597} & 0.643 \\
\multicolumn{1}{l}{} & RF & 0.710 & 0.676 & 0.667 & 0.635 & 0.636 & \textbf{0.605} & & 0.596 & 0.690 & 0.739 & 0.601 & 0.601 & \textbf{0.642} \\
 & Baseline & 0.247 & 0.316 & 0.379 & 0.383 & 0.376 & 0.385 & & 1.545 & 1.456 & 1.379 & 1.017 & 1.029 & 0.998 \\\bottomrule
\end{tabular}\label{tab_prediction_2}
\end{table*}

\subsection{Treatment Effect Estimation}\label{sub:effect}
With the 6 GBDT models trained (using Q4 of 2018 to predict months 1-6 of 2019), we are able to make counterfactual predictions of the outcomes of the target repositories in months 1-6 of 2020 (based on their features in Q4 of 2019). Note that repositories with missing features are dropped, which leaves us with 48,612 repositories in the target dataset. The effect of the pandemic in a given month (1-6) is then measured by the difference between the observed and the counterfactual outcomes.

\begin{figure}[htb]
    \centering
    \begin{center}
        \subfigure[Productivity (Pushes). \label{fig:res_push}]
        {\includegraphics[width=0.85\linewidth]{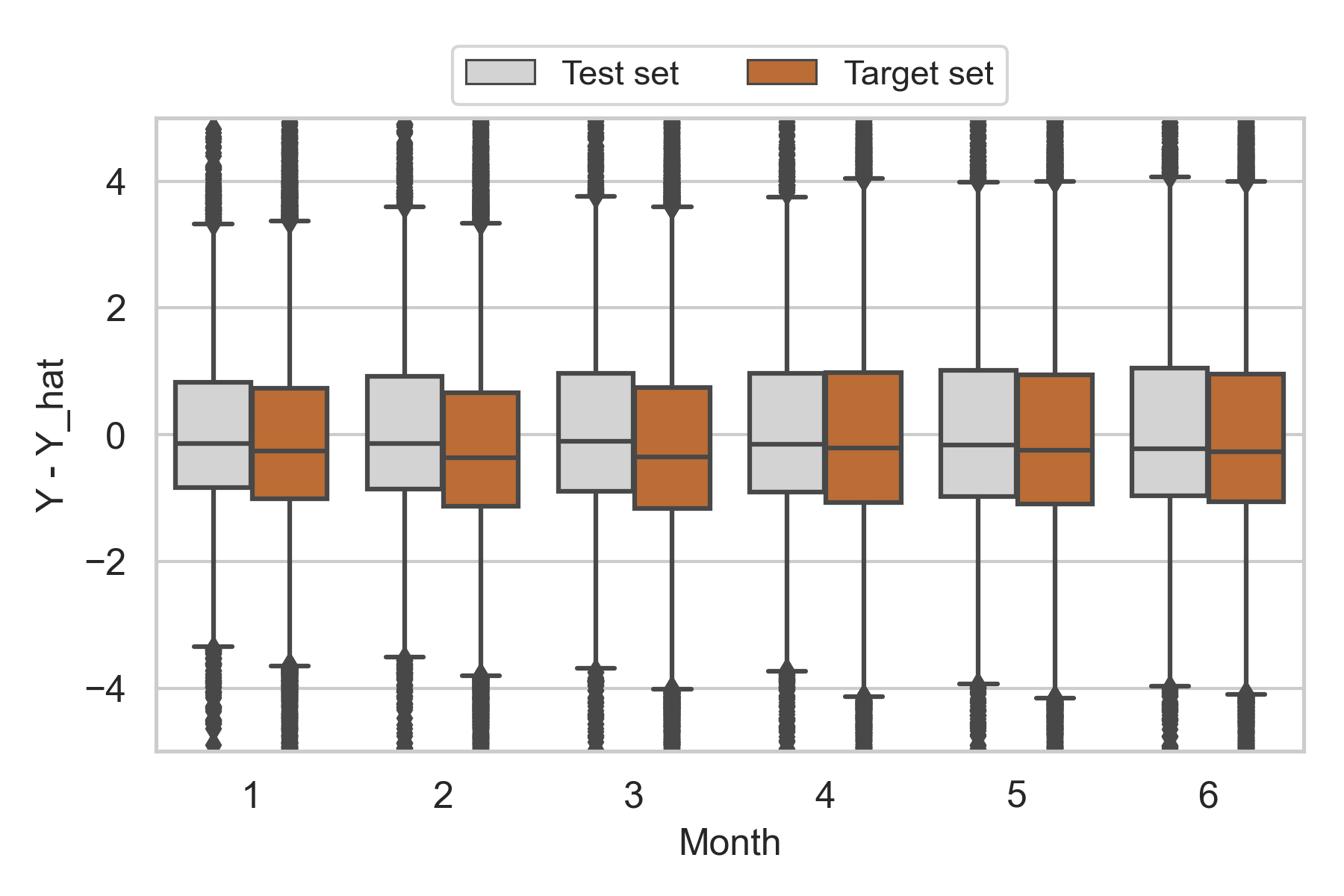}}
        \subfigure[Growth (Active Team Size). \label{fig:res_size}]
        {\includegraphics[width=0.85\linewidth]{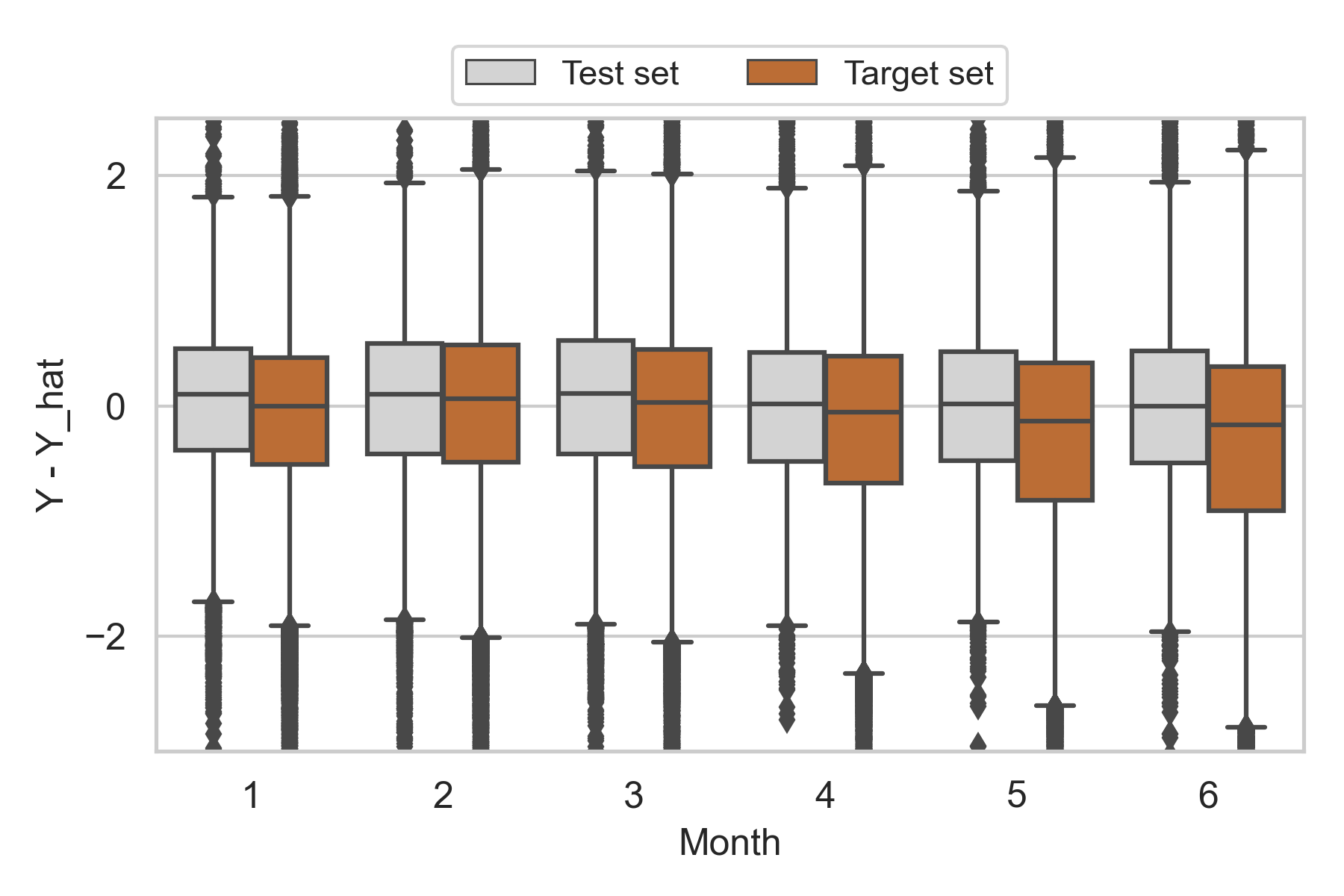}}
        \caption{Distribution of ITE (of the target set): the pandemic has an immediate effect on team productivity and a delayed effect on team size. Residuals of predictions (of the reference-test set) are centered closely to zero. }\label{fig:res}
    \end{center}
\end{figure} 

We plot the distributions of the individual treatment effects in Figure~\ref{fig:res}, along the time horizons. The mean of the distribution in a particular month is essentially the average treatment effect of the pandemic in 1-6 months on the selected teams (teams that are already active before the pandemic). When the mean is below zero, it means the productivity or size of the teams has declined because of the pandemic. Comparing to Figure~\ref{fig:sub_push_12} and ~\ref{fig:sub_actor_12}, the average treatment effects on these selected teams are much more stable over months. As a reference, we also show the residuals between the observed and predicted outcomes in 2019 (calculated based on the 20\% test data). These residuals indicate the individual treatment effects of ``no pandemic'', and therefore they should be centered at zero if the machine learning predictors are unbiased. As we can see from Figure~\ref{fig:res}, when there was no pandemic, there was indeed no obvious difference between the observed/predicted team outcomes. When the pandemic hits, there are noticeable negative effects on team productivity, especially in the first three months of 2020. Interestingly, these negative effects are not clear on team size in the same months (1-3), while from April (4), we start to observe a clearly declining pattern (in team size), especially in May (5) and June (6). For each month, we ran Two-sample Kolmogorov-Smirnov tests and the null hypothesis (``the pre- and post-pandemic distributions are identical'') is rejected for all settings in Figure~\ref{fig:res}.  

It is intriguing to observe that team productivity reduced immediately following the shock (1-3) and then recovered to normal a few months after (4-6). The number of active team members, on the other hand, did not immediately drop when the pandemic hit (1-3), but it drops gradually when the teams respond in a longer term (5-6). A possible explanation is that the core members of a team kept contributing (and even took more responsibilities in a longer term), while the activeness of peripheral members are more likely to be impacted by the pandemic - and some of them may not come back.

Although the average effects are now more stable over time, we still observe a high variance in treatment effects on individual teams. This motivates us to further analyze what kind of teams are more likely to be affected by the pandemic and what teams are more likely to bounce back quickly. 

\subsection{Prediction Error Inference}\label{sub:conformal}

Are the counterfactual prediction models trustworthy for the downstream analysis of heterogeneous treatment effects, especially when the prediction task is quite difficult? The key to answer this question is to understand the effects of the potential prediction errors. In other words, if we can derive findings that are robust to the potential errors of the counterfactual predictions, the analysis is not limited by the goodness of the prediction models. 

The challenge is how to estimate the prediction errors. Recall that all teams are ``treated'' during the pandemic, we can not directly obtain the prediction error because they are counterfactual. We adopt a principled statistical method, \textit{split conformal prediction}~\cite{lei2018distribution}, to infer the errors of the out-of-sample counterfactual predictions from the distribution of residuals on the test data.\footnote{The conformal inference can be made under the assumption of exchangeability of the two datasets, which holds as we can predict future outcomes for teams in the target set with models trained with the reference set.}

More specifically, given a model trained for a specific month, denote the prediction error by $E$ and the number of repositories in the test set by $n$ (i.e., 20\% of the reference set), the goal is to obtain a conformal interval of $E$ given a miscoverage level $\alpha$ ($\alpha=5\%$). We infer the conformal interval of $E$ in the target set with the following two steps (see Algorithm 2 in~\cite{lei2018distribution}). (1) We draw the residuals of predictions for each repository in the test set and get their absolute values. (2) We rank the absolute residuals and get the $k$-th smallest value $d$, where $k=\lceil(n+1)(1-\alpha)\rceil$. The probability that $E$ falls within $[-d, d]$ is no less than $1-\alpha$ (i.e., 95\%). This conformal interval gives us a reasonable estimation of the error distribution of the counterfactual predictions and can be used for resampling residuals for bootstrapped regressions in the following analysis.
\section{Heterogeneous Effect Analysis}\label{sec:hete}
The high variance in the individual treatment effects in Figure~\ref{fig:res} shows the heterogeneity of teams in reacting to the shock, which can be related to the inherent properties of teams. The capacity of a team to cope and recover from adversity is known as \textit{team resilience}~\cite{alliger2015team}. To understand what kind of teams are more resilient/prone to the shock, we can relate the individual treatment effects to team properties with regressions to investigate how the properties can explain the heterogeneity in the effects. 

In specific, we select repositories in the target set for the regressions and use the same properties (in Q4 of 2019) in the preceding analysis to describe them. We perform bootstrapped regressions with a strategy similar to residual resampling~\cite{davison2002introduction}, yet the ``residuals'' are drawn from the generated conformal interval $[-d, d]$. Particularly, for each iteration of the bootstrap, we generate an infused dependent variable with a noise $\epsilon$ randomly drawn from $[-d, d]$,\footnote{We keep the empirical distribution of residuals in the test set and randomly sample residuals from it. Only values between $-d$ and $d$ are kept.} regress it on team properties, and repeat this process 1,000 times to obtain a median and a 95\% confidence interval for coefficients of each independent variable. In this way, for independent variables whose coefficient values in the 95\% confidence interval are on the same side of zero (i.e., zero is not contained in the interval), we can safely derive conclusions about their positive/negative relations to the dependent variable as the coefficient can be considered significant. Independent variables with positive coefficients contribute positively to team resilience.

\begin{table}[h]
\small
\centering
\caption{Bootstrapped OLS regressions of shock effects (on team productivity) in March 2020. Independent variables with coefficients significantly above or below zero are in bold; the corresponding median values are marked with *.}
\begin{tabular}{lll}
\toprule
\multicolumn{3}{c}{Dependent Variable: $y-(\hat{y}+\epsilon)$, $y=log(pushes+1)$}                \\\midrule
 & Median & 95\% CI  \\\midrule
\textbf{const} & 6.7E-01* & [3.3E-01, 9.7E-01]\\
log(members+1) & 1.6E-02 & [-1.1E-02, 4.1E-02]\\
\textbf{log(dedicated members+1)} & -1.7E-02* & [-3.2E-02, -1.6E-03]\\
\textbf{Avg. contributed repositories} & 3.3E-03* & [9.1E-04, 5.7E-03]\\
Max. of platform tenures & -2.0E-05 & [-5.0E-05, 6.9E-06]\\
Med. of platform tenures & -2.3E-05 & [-6.0E-05, 1.3E-05]\\
\textbf{CV of platform tenures} & -1.6E-03* & [-2.5E-03, -6.6E-04]\\
\textbf{Max. of coding tenures} & 2.8E-04* & [1.7E-04, 3.8E-04]\\
\textbf{Med. of coding tenures} & -2.8E-04* & [-3.3E-04, -2.2E-04]\\
CV of coding tenures & -4.3E-05 & [-9.2E-04, 8.6E-04]\\
Med. of team tenures & 3.0E-05 & [-9.9E-06, 7.5E-05]\\
\textbf{Max. of log(followers+1)} & -8.9E-03* & [-1.6E-02, -2.2E-03]\\
Med. of log(followers+1) & 8.1E-03 & [-2.0E-02, 3.7E-02]\\
CV of log(followers+1) & 3.1E-05 & [-9.5E-05, 1.7E-04]\\
\textbf{Unique countries} & -1.7E-02* & [-2.4E-02, -9.7E-03]\\
Unique program. lang. & 3.7E-03 & [-2.3E-03, 1.0E-02]\\
Len. off segment (TH=16) & -1.1E-03 & [-5.3E-03, 3.0E-03]\\
Len. off segment (TH=32) & -2.3E-03 & [-8.7E-03, 3.9E-03]\\
\textbf{Len. off segment (TH=64)} & -1.8E-02* & [-2.9E-02, -5.9E-03]\\
\textbf{Entropy of working days} & -2.0E-01* & [-2.6E-01, -1.5E-01]\\
Prop. emoji posts & 9.6E-02 & [-7.5E-03, 2.1E-01]\\
Avg. monthly push, log scale & -1.1E-02 & [-2.3E-02, 2.1E-03]\\
\textbf{log(pushes+1)} & 2.1E-02* & [1.3E-02, 3.0E-02]\\
\textbf{log(issues+1)} & 1.5E-02* & [4.9E-03, 2.4E-02]\\
\textbf{log(pull requests+1)} & -1.0E-02* & [-2.0E-02, -1.6E-03]\\
\textbf{log(all watches+1)} & -1.3E-02* & [-1.7E-02, -7.7E-03]\\
Days since created & -4.1E-07 & [-1.6E-05, 1.6E-05]\\\bottomrule                
\end{tabular}
\label{tab:regression_push}
\end{table}
\begin{table}[h]
\small
\centering
\caption{Bootstrapped OLS regressions of shock effects (on team growth) in June 2020. Independent variables with coefficients significantly above or below zero are in bold; the corresponding median values are marked with *.}
\begin{tabular}{lll}
\toprule
\multicolumn{3}{c}{Dependent Variable: $y-(\hat{y}+\epsilon)$, $y=log(members+1)$}                \\\midrule
 & Median & 95\% CI \\\midrule
\textbf{const} & -2.2E+00* & [-2.3E+00, -2.0E+00]\\
\textbf{log(members+1)} & 2.4E-01* & [2.2E-01, 2.5E-01]\\
\textbf{log(dedicated members+1)} & -3.7E-02* & [-4.5E-02, -2.9E-02]\\
Avg. contributed repositories & -6.4E-04 & [-1.9E-03, 6.5E-04]\\
\textbf{Max. of platform tenures} & -2.0E-05* & [-3.5E-05, -4.4E-06]\\
\textbf{Med. of platform tenures} & 3.7E-05* & [1.8E-05, 5.6E-05]\\
CV of platform tenures & 2.2E-04 & [-2.8E-04, 6.6E-04]\\
Max. of coding tenures & 1.3E-05 & [-3.9E-05, 6.8E-05]\\
\textbf{Med. of coding tenures} & 6.0E-05* & [3.2E-05, 8.8E-05]\\
CV of coding tenures & 2.5E-04 & [-1.9E-04, 6.6E-04]\\
\textbf{Med. of team tenures} & -7.2E-05* & [-9.4E-05, -4.8E-05]\\
\textbf{Max. of log(followers+1)} & 8.8E-03* & [5.5E-03, 1.2E-02]\\
\textbf{Med. of log(followers+1)} & -6.1E-02* & [-7.4E-02, -4.5E-02]\\
\textbf{CV of log(followers+1)} & -1.1E-04* & [-1.8E-04, -4.5E-05]\\
\textbf{Unique countries} & -3.0E-02* & [-3.4E-02, -2.6E-02]\\
\textbf{Unique program. lang.} & -1.4E-02* & [-1.6E-02, -1.0E-02]\\
\textbf{Len. off segment (TH=16)} & 1.1E-02* & [9.3E-03, 1.4E-02]\\
Len. off segment (TH=32) & 8.0E-04 & [-2.4E-03, 4.2E-03]\\
\textbf{Len. off segment (TH=64)} & -9.1E-03* & [-1.5E-02, -3.3E-03]\\
\textbf{Entropy of working days} & 4.7E-01* & [4.4E-01, 5.0E-01]\\
Prop. emoji posts & -5.0E-02 & [-1.0E-01, 5.7E-03]\\
Avg. monthly push, log scale & -3.8E-03 & [-1.0E-02, 2.4E-03]\\
\textbf{log(pushes+1)} & 2.1E-02* & [1.7E-02, 2.5E-02]\\
\textbf{log(issues+1)} & 2.3E-02* & [1.8E-02, 2.8E-02]\\
\textbf{log(pull requests+1)} & 1.2E-02* & [7.6E-03, 1.7E-02]\\
log(all watches+1) & 2.4E-03 & [-2.0E-04, 4.9E-03]\\
\textbf{Days since created} & 8.6E-06* & [5.5E-07, 1.6E-05]\\\bottomrule        
\end{tabular}
\label{tab:regression_size}
\end{table}

Adding all features into a regression model could lead to collinearity and deteriorate the results. To address this problem, we remove features that are highly collinear with others. Specifically, we run a hierarchical clustering and select one feature for each cluster in which the pairwise Spearman correlation coefficients are all higher than 0.7~\cite{dormann2013collinearity}. In this way, we obtain 26 features from the 45 features in total. The variance inflation factor (VIF) of the selected features are all $<$10: only one (\textit{log(members+1)}) has VIF=9.6 and all others $<$5, indicating multicollinearity is under control~\cite{dormann2013collinearity}.

With the selected features as independent variables, we perform OLS regressions for the ITEs of each of the six months in 2020. The independent variables with long-tailed distributions have been converted to log-scales to reduce the effect of outliers. 

\subsection{Heterogeneous Effects on Team Productivity and Growth}

\begin{figure}
\begin{center}
    \includegraphics[width=0.85\linewidth]{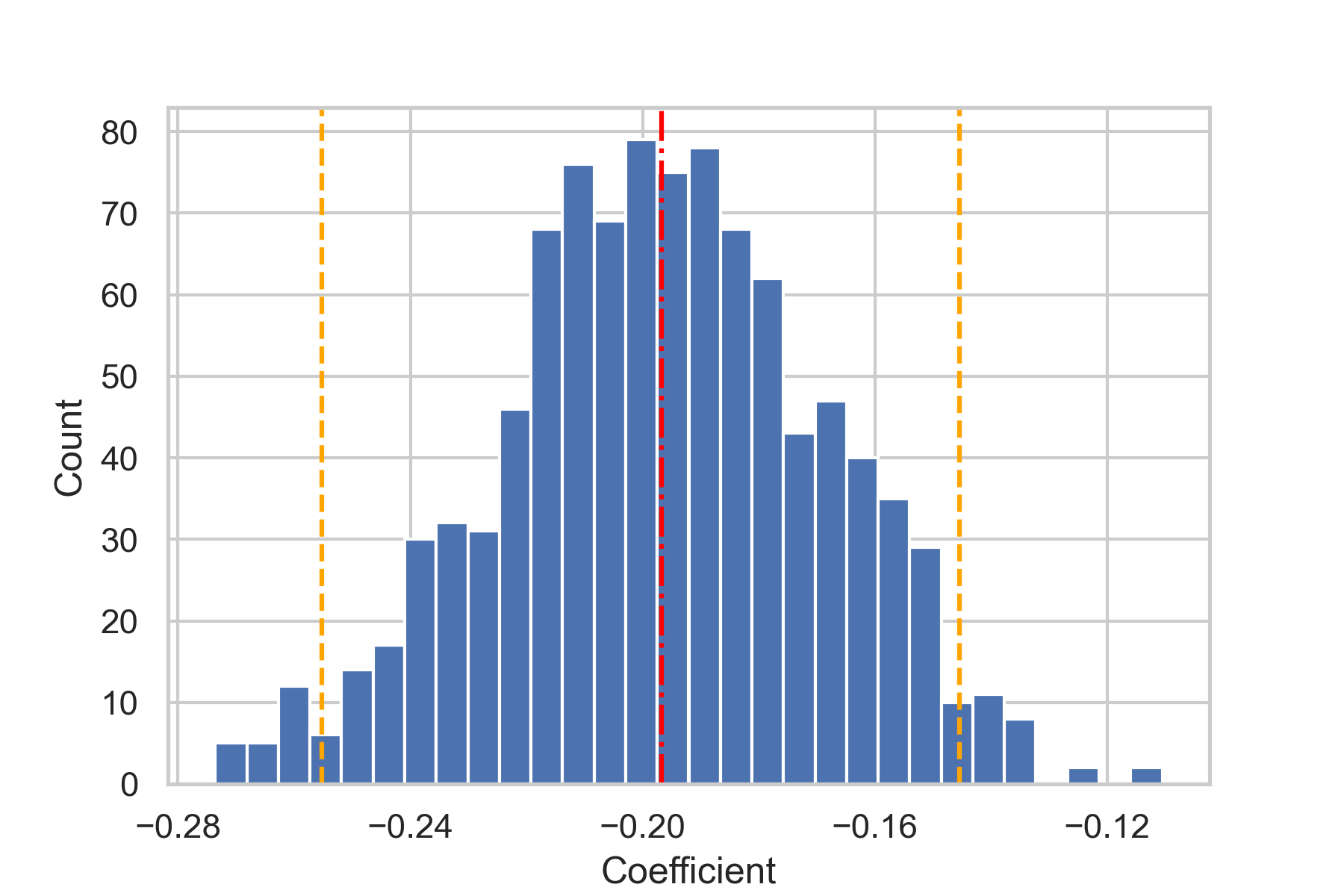}
    \caption{Distribution of the coefficient of \textit{entropy of working days} in bootstrapped regressions (outcome variable: shock effect on productivity in March 2020). Red dashdot line shows the median of the coefficient. Orange dashed lines show 95\% confidence interval of the coefficient. The variable has a significantly negative effect on the outcome. } 
    \label{fig:push_3_week_entropy}
    \end{center}
\end{figure}

We report the bootstrapped regressions on productivity in March 2020 and that on team size in June 2020, in which we observe obvious shock effects (Figure~\ref{fig:res}). 
Specifically, we report the median and the 95\% confidence interval of the coefficient for each independent variable in these two settings. We illustrate the significance of coefficients with an example in Figure~\ref{fig:push_3_week_entropy}. When the dependent variable is the shock effects on team productivity in March 2020, the coefficients of the variable \textit{entropy of working days} in the 1,000 bootstrapped regressions have a median of $-0.2$ and a 95\% confidence interval of $[-0.26, -0.15]$, showing that the negative relation between that variable and the outcome is significant. 

We then examine the effect of certain variables more closely from the tables and note the following observations. 

\underline{Team size}: larger teams are more resilient to the shock in terms of maintaining their sizes/growth. Size doesn't show a significant relation to maintaining productivity. 

\underline{Multitasking}: Teams with a higher number of dedicated members (active only in this repository) are less resilient to the shock in terms of both productivity and growth. A high proportion (98\% in~\cite{vasilescu2016sky}) of prolific developers contribute to multiple projects. With productivity levels controlled, a high number of dedicated members may indicate the irreplaceability of their roles, which reduces the flexibility of the team under shock. On the other hand, members contributing to more repositories at the same period could mean a higher level of engagement, which makes the team more resilient in terms of productivity (but not growth).

\underline{Tenure}: The resilience in productivity increases when the max coding tenure increases, indicating the benefit to have senior contributors/experts. The median of coding tenures, however, shows a negative correlation with the productivity resilience. One possible explanation is that the senior developers are more likely playing roles of management and coordination in the team (just like senior developers in IT companies), while junior developers may spend more time to learn~\cite{lee2013github} and make contributions to build their reputation~\cite{lakhani2004open}. It is not surprising to observe a positive correlation between the median of coding tenures and the team size resilience, as teams with less junior developers may have more room for junior developers to join. Teams that have more senior people and more new members are more resilient in maintaining their sizes, while the latter may help recruit new members and the former may help retain existing members. The max platform tenure shows a negative effect on team size resilience. Diversity in platform tenures shows negative effect on productivity resilience, possibly due to different levels of adaptability to the remote collaboration mode.

\underline{Prestige}: Member prestige (measured by the number of followers) has no significant effect on productivity resilience except for the max number of followers. The max number of followers of a team member is positively related to resilience in growth, possibly because that popular users help attract their followers to new projects~\cite{blincoe2016understanding}. However, both the median and the diversity of prestige show a negative correlation with the resilience in growth, suggesting that even if a team has some very popular members, having a large number of junior members still helps attract new members. The new members attracted by the most popular member may not be able to immediately contribute to productivity, which helps explain the negative correlation between the max number of followers and productivity resilience.

\underline{Activity}: Teams with a higher level of activity (\textit{push} or \textit{issues}) are more resilient in both productivity and growth. Teams with a higher level of \textit{pull requests} are less resilient in productivity while more resilient in growth, probably because a higher level of collaboration helps retention. Teams with more \textit{watchers} are less likely to maintain their productivity under the shock.  

\underline{Diversity}: Open source software development often benefits from worldwide collaborations. However, our results show that the country diversity plays a negative role in resilience in both outcomes under the shock. Teams with members from more countries are more likely to experience a reduction in productivity and size, possibly because teams with members from many countries tend to be loosely connected~\cite{wuthnow1998loose}, or because the policies and contexts in different countries add more uncertainties and complications to the challenges under a worldwide pandemic.

The programming languages that one masters represent their technical background. Programming language diversity has a negative effect on growth, likely because it is hard to recruit members with similar technical backgrounds. 

\underline{Work schedule}: The length of off segment with a high threshold (64) is negatively correlated with both productivity and team size resilience. Off segment with a low threshold (16) is positively related to resilience in growth. A longer off-segment means fewer members work in these hours. At a higher cutoff threshold, this can be due to different time zones of members or a spanned-out working schedule. At a lower threshold, this indicates most members maintain a clear boundary between work and life, which is beneficial to the growth of a team. The entropy of working days in a week shows a negative effect on productivity resilience, which implies that a spread-out work schedule doesn't help maintain productivity under shock. This however has a positive relation to the resilience in team size.

\underline{Others}: Team age increases resilience in growth but not in productivity.
The proportion of emoji posts has no significant effect, which may imply that a more accurate way of measuring the emotional status of teams is needed. 

\subsection{Discussion}

We then examine the consistency of the significance and directions for the coefficients of team properties in multiple regressions across the six months (see Appendix). Note that the effect of the shock on a team can be both short-term and longer-term, and the signs and significance levels of corresponding factors may vary across different time horizons. 

Most features present a consistent sign (but some with various coefficients and significance levels) over the time horizons. Some present interesting patterns. The number of countries, the entropy of working days, activity levels in \textit{push} and \textit{issues} are fairly robust across time horizons for productivity resilience.  Team size, the median of platform tenure, number of dedicated members, member prestige, number of countries, number of programming languages, and off segments are fairly robust for resilience in team size. The proportion of emoji posts has a significantly positive effect on team size resilience in the first four months (months 1, 2, and 4 for productivity), but the effect fades in later months.  This indicates that emotional intelligence may be an effective factor to resist the immediate effects of a shock, but its effect might not last long enough. Team productivity levels before the pandemic have a negative effect on the resilience of team size in the first three months but the effect becomes positive in later months. Highly productive teams might be busy maintaining their activity levels in a short term, but when the situation stables, they will be back on track for quick growth. Older teams seem to be more resilient overall, although the effect size is small. 

\section{Implications}

Repositories on GitHub are featured in cross-continent and cross-cultural collaborations. However, our analysis show that teams with a higher level of country diversity are more vulnerable to the COVID-19 shock in both productivity and growth. This implies that teams with high diversity of countries should consider to establish or strengthen team bonds as well as increasing the substitutability of the members to increase their resilience under shock. 

The social influence of the popular developers~\cite{blincoe2016understanding} still exists during the shock, as the maximum of followers shows a positive coefficient across most of the regressions of growth. Meanwhile, the role of senior developers (with maximum coding tenure) is salient in productivity resilience. Junior developers and new team members also play important roles in the resilience in productivity or growth. This indicates that a robust team should carefully invest in a healthy balance between experience and energy. 

The multitasking level shows positive correlations with both the team size and productivity resilience, declaring that teams with more multitasking members are less affected by the shock. This finding implies that instead of relying on ``full-time'' members, getting more developers who contribute partially engaged could still benefit the team resilience. This could be achieved by being open to collaborations with other teams or recruiting part-time members. 

\section{Limitations}
There are some limitations in our analysis. First, gender diversity is considered as an important property of GitHub teams~\cite{Vasilescu2015-ql} but is not included in our feature set. Gender information of developers is often inferred from their country and real name in literature~\cite{Vasilescu2015-ql, ortu2017diverse}. However, the real names are not accessible in our dataset\footnote{https://ghtorrent.org/faq.html, retrieved on 10/20/2021.} due to GDPR. Even so, we argue that the absence of gender diversity does not affect the validity of this analysis as the confounds are carefully controlled. Second, we focus on one metric (i.e., \textit{pushes}) when investigating the shock effects on productivity, thus the conclusions may not be applicable to other metrics. Finally, this analysis is conducted on the online collaborative platform of developers in public repositories. Our findings are directly applicable to OSS communities and we anticipate they be generalized to related contexts (e.g., IT companies, research teams, or non-profit organizations) to certain extent, given the frequent connections of these contexts to GitHub in literature~\cite{kalliamvakou2015open, perkel2016democratic, kummer2020unemployment}. However, generalization to other organizational contexts should be verified with caution. The method developed in this study is general and we recommend the readers to reproduce our analysis and compare the findings in other industry organizations when team properties and team outcomes before and after shock are available.

\section{Conclusion}

In this paper, we study the effects of the COVID-19 pandemic shock on GitHub repositories (as surrogates of remote teams) during the early months. We find that under the shock, the GitHub community show different patterns of response in terms of different metrics. In the level of individual teams, the effect of the shock can be immediately observed in team productivity, while the effect on team growth is lagged. Team properties such as the country diversity, multitasking level, member experience and prestige, and emotions, are revealed as correlated with the resilience of an individual team under the shock.

\section{Potential broader impact and ethics considerations}

Results from this analysis could help organizations and individuals to understand the resilience of remote teams under shock and could potentially help them make decisions about remote work post-pandemic and on a longer horizon. Data used in this study are from third-party platforms (i.e., GHArchive and GHTorrent). We do not collect or release a new dataset. During data processing, we do not violate ethical principles or attempt to link the identities of developers. 

\section{Acknowledgement}

The authors would like to thank the anonymous reviewers of the previous version for their insightful comments, which have inspired our new analysis on the inference of errors of the counterfactual predictions and robust regressions of ITE.

\bibliography{github_team}

\appendix
\section{Appendix}

\begin{table*}[h]
\footnotesize
\centering
\caption{\footnotesize Bootstrapped OLS regressions of shock effects on team productivity (Top table) and team size (Bottom table). The median values of coefficients are reported and are marked with * if they are significantly above or below zero.}
\begin{center}
\begin{tabular}[t]{lcccccc}					
\toprule
\multicolumn{7}{c}{Dependent Variable: $y-(\hat{y}+\epsilon)$, $y=log(pushes+1)$}\\\hline
	& 1 & 2 & 3 & 4 & 5 & 6	\\					
\midrule							
const& -2.1E-01& 7.8E-02& 6.7E-01*& 2.4E-01& 1.7E-01& 1.7E-01\\
log(members+1)& -8.3E-03& 2.7E-02*& 1.6E-02& -1.2E-02& -1.5E-02& 3.1E-02*\\
log(dedicated members+1)& -4.5E-03& -1.9E-03& -1.7E-02*& -1.6E-02& 9.3E-04& -1.7E-02*\\
Avg. contributed repositories& 1.5E-03& 2.0E-03& 3.3E-03*& 8.4E-04& 6.1E-03*& 6.0E-03*\\
Max. of platform tenures& 5.1E-05*& -1.1E-05& -2.0E-05& 7.0E-06& -6.6E-06& 1.0E-05\\
Med. of platform tenures& -2.3E-05& 2.2E-06& -2.3E-05& -1.4E-05& 3.0E-06& 1.4E-05\\
CV of platform tenures& -1.3E-03*& -6.4E-04& -1.6E-03*& -1.3E-03*& -1.5E-03*& -7.9E-05\\
Max. of coding tenures& 1.4E-04*& 1.8E-04*& 2.8E-04*& 1.6E-04*& 7.5E-05& 5.6E-05\\
Med. of coding tenures& -5.4E-05*& -8.4E-05*& -2.8E-04*& -3.5E-05& 1.4E-05& -4.2E-05\\
CV of coding tenures& -3.9E-04& 1.2E-03*& -4.3E-05& 9.0E-04& 6.6E-04& 1.4E-03*\\
Med. of team tenures& -4.4E-05*& -2.5E-05& 3.0E-05& 1.2E-05& -7.0E-05*& 3.0E-05\\
Max. of log(followers+1)& -7.0E-03*& -6.0E-03& -8.9E-03*& 4.6E-03& 6.6E-03& 6.4E-03\\
Med. of log(followers+1)& -5.8E-03& -1.7E-02& 8.1E-03& -2.2E-02& -1.3E-02& 3.5E-03\\
CV of log(followers+1)& 1.1E-04& -4.5E-06& 3.1E-05& 3.0E-05& 4.2E-05& -1.9E-04*\\
Unique countries& -4.4E-03& -1.3E-02*& -1.7E-02*& -1.6E-02*& -1.3E-02*& -1.8E-02*\\
Unique programming languages& -1.3E-03& 7.5E-03*& 3.7E-03& 1.1E-02*& 4.0E-03& 5.0E-03\\
Length of off segment (TH=16)& -8.8E-04& 3.2E-03& -1.1E-03& -5.1E-03*& -4.5E-03& 3.4E-04\\
Length of off segment (TH=32)& 8.4E-03*& 1.4E-03& -2.3E-03& -6.8E-04& -2.6E-03& -8.7E-03*\\
Length of off segment (TH=64)& -5.1E-03& -1.9E-02*& -1.8E-02*& -1.4E-02*& -1.4E-02*& -6.3E-03\\
Entropy of working days& -1.3E-01*& -8.8E-02*& -2.0E-01*& -3.5E-02& 6.1E-02*& -1.1E-01*\\
Prop. emoji posts& 1.7E-01*& 1.5E-01*& 9.6E-02& 1.4E-01*& 7.0E-02& -9.8E-02\\
Avg. monthly push, log scale& -4.0E-03& -1.5E-02*& -1.1E-02& -1.0E-02& -1.3E-02*& -1.8E-02*\\
log(pushes+1)& 2.8E-02*& 1.6E-02*& 2.1E-02*& 1.2E-02*& 3.1E-04& -3.9E-04\\
log(issues+1)& 1.6E-02*& 2.6E-03& 1.5E-02*& 1.8E-02*& 2.3E-02*& 2.7E-02*\\
log(pull requests+1)& -9.0E-04& -2.2E-02*& -1.0E-02*& -1.9E-02*& -1.3E-02*& -6.0E-03\\
log(all watches+1)& -1.3E-02*& -1.2E-02*& -1.3E-02*& -9.2E-04& 3.9E-03& -1.0E-02*\\
Days since created& 3.9E-05*& 3.4E-05*& -4.1E-07& 3.0E-05*& 4.6E-05*& 7.9E-05*\\
\bottomrule							
\end{tabular}
\label{tab:table1_a}
\end{center}
\begin{center}
\begin{tabular}{lcccccc}							
\toprule
\multicolumn{7}{c}{Dependent Variable: $y-(\hat{y}+\epsilon)$, $y=log(members+1)$}\\\hline
	& 1 & 2 & 3 & 4 & 5 & 6	\\					
\midrule							
const& -1.3E-01& -4.1E-01*& -2.0E-02& -2.3E+00*& -2.0E+00*& -2.2E+00*\\
log(members+1)& 1.1E-01*& 1.3E-01*& 1.1E-01*& 2.3E-01*& 2.2E-01*& 2.4E-01*\\
log(dedicated members+1)& -2.5E-02*& -3.4E-02*& -2.2E-02*& -3.9E-02*& -2.9E-02*& -3.7E-02*\\
Avg. contributed repositories& 1.0E-03& 2.7E-03*& 2.0E-03*& -2.9E-04& 2.5E-03*& -6.4E-04\\
Max. of platform tenures& 2.9E-05*& -6.2E-05*& -3.8E-05*& -1.7E-05*& -1.3E-05& -2.0E-05*\\
Med. of platform tenures& 2.3E-05*& 7.5E-05*& 3.2E-05*& 5.4E-05*& 2.4E-05*& 3.7E-05*\\
CV of platform tenures& -3.6E-04& 6.8E-04*& 8.0E-04*& 5.7E-04*& -6.7E-04*& 2.2E-04\\
Max. of coding tenures& -1.0E-04*& -1.1E-04*& -1.1E-04*& 9.8E-05*& 1.4E-04*& 1.3E-05\\
Med. of coding tenures& 6.4E-05*& 1.4E-04*& 6.0E-05*& 8.0E-05*& -1.7E-04*& 6.0E-05*\\
CV of coding tenures& 7.6E-04*& 1.6E-03*& 8.7E-04*& 1.6E-03*& -1.2E-03*& 2.5E-04\\
Med. of team tenures& 2.7E-05*& 4.4E-05*& 4.9E-05*& 1.4E-05& -4.3E-06& -7.2E-05*\\
Max. of log(followers+1)& 9.5E-03*& 6.6E-03*& 3.9E-03*& 9.7E-03*& 1.1E-02*& 8.8E-03*\\
Med. of log(followers+1)& -4.5E-03& -1.9E-02*& 6.1E-05& -4.8E-02*& -3.5E-02*& -6.1E-02*\\
CV of log(followers+1)& -2.3E-04*& -2.0E-04*& -1.8E-04*& -2.7E-04*& -8.3E-05*& -1.1E-04*\\
Unique countries& -7.8E-03*& -1.1E-02*& -1.2E-02*& -2.3E-02*& -3.0E-02*& -3.0E-02*\\
Unique programming languages& -1.3E-02*& -7.8E-03*& -8.4E-03*& -1.8E-02*& -8.8E-03*& -1.4E-02*\\
Length of off segment (TH=16)& 1.5E-03& 5.2E-03*& 2.3E-03& 9.3E-03*& 8.5E-03*& 1.1E-02*\\
Length of off segment (TH=32)& -3.5E-03& -8.2E-04& -5.3E-04& -7.1E-04& 1.2E-03& 8.0E-04\\
Length of off segment (TH=64)& -9.5E-03*& -9.8E-03*& -1.1E-02*& -5.9E-03*& -1.1E-02*& -9.1E-03*\\
Entropy of working days& -4.0E-02*& -2.2E-02& -5.9E-02*& 4.4E-01*& 4.8E-01*& 4.7E-01*\\
Prop. emoji posts& 1.5E-01*& 1.4E-01*& 1.6E-01*& 1.1E-01*& 4.4E-02& -5.0E-02\\
Avg. monthly push, log scale& 5.2E-03& 1.4E-02*& 6.2E-03& -2.5E-03& -1.0E-02*& -3.8E-03\\
log(pushes+1)& -1.1E-02*& -9.8E-03*& -1.3E-02*& 1.2E-02*& 1.5E-02*& 2.1E-02*\\
log(issues+1)& -3.6E-03& -4.6E-04& 5.5E-03*& 1.8E-02*& 2.9E-02*& 2.3E-02*\\
log(pull requests+1)& 3.2E-03& 5.7E-03*& 9.2E-03*& -4.3E-03& -3.9E-03& 1.2E-02*\\
log(all watches+1)& -1.8E-02*& -1.1E-02*& -1.8E-02*& 1.2E-02*& 9.6E-03*& 2.4E-03\\
Days since created& 3.1E-05*& 3.6E-05*& 2.5E-05*& 3.7E-05*& 3.9E-05*& 8.6E-06*\\
\bottomrule							
\end{tabular}
\label{tab:table1_b}
\end{center}
\label{tab:appendix}

\end{table*}

\end{document}